\documentclass[10pt,twocolumn,letterpaper]{article}

\usepackage{iccv}
\usepackage{times}
\usepackage{epsfig}
\usepackage{graphicx}
\usepackage{amsmath}
\usepackage{amssymb}
\usepackage{multirow}

\usepackage[pagebackref=true,breaklinks=true,letterpaper=true,colorlinks,bookmarks=false]{hyperref}

\iccvfinalcopy 

\def\iccvPaperID{4362} 

\ificcvfinal\fi

\begin{document}

\definecolor{mgreen}{rgb}{0, 0.5, 0}
\definecolor{mred}{rgb}{0.8, 0, 0}
\definecolor{mblue}{rgb}{0, 0, 0.8}
\definecolor{morange}{rgb}{0.85, 0.5, 0}

\title{C$^{4}$Net: Contextual Compression and Complementary Combination \\ Network for Salient Object Detection}

\author{Hazarapet Tunanyan\\
{\small Picsart AI Research  (PAIR)}\\
{\tt\small hazarapet.tunanyan@picsart.com}
}

\maketitle
\ificcvfinal\thispagestyle{empty}\fi

\begin{abstract}
    Deep learning solutions of the salient object detection
    problem have achieved great results in recent years. The
    majority of these models are based on encoders and decoders, with a different multi-feature combination. In this
    paper, we show that feature concatenation works better
    than other combination methods like multiplication or addition. Also, joint feature learning gives better results, because of the information sharing during their processing.
    We designed a Complementary Extraction Module (CEM)
    to extract necessary features with edge preservation. Our
    proposed Excessiveness Loss (EL) function helps to reduce
    false-positive predictions and purifies the edges with other
    weighted loss functions. Our designed Pyramid-Semantic
    Module (PSM) with Global guiding flow (G) makes the prediction more accurate by providing high-level complementary information to shallower layers. Experimental results
    show that the proposed model outperforms the state-of-the-art methods on all benchmark datasets under three evaluation metrics.
\end{abstract}

\section{Introduction}
\label{sec:intro}
Salient object detection (SOD) aims to detect the most visually attractive parts of images and videos, which is widely used in many applications like visual tracking \cite{sal_track}, image retrieval \cite{obj_retr}, content-aware image
editing \cite{RepFinder}, robot navigation \cite{robot_nav}, and many others. For many years, researchers have proposed some solutions \cite{old_sod1, old_sod2, old_sod3, old_sod4}, which are based on hand-crafted features (\eg color, contrast, and texture), however, these approaches can not take into account high-level semantic information, which is a big restriction for the solution of the problem. The recent development of deep convolutional neural networks (CNNs) demonstrated powerful capabilities in terms of the high (\eg class, position) and low (\eg high-frequency data, detailed information) level feature extraction, which promotes better results of the salient object detection problem. \\
Despite good performance achievements, there is still a large place for improvements. As many modern solutions propose U-Net-like architectures to solve the problem, the feature combination and their processing methods have big potential to improve. Many models observe feature fusing methods with different aggregation functions like multiplication and addition. We find these methods have some drawbacks, because of the choice of aggregation functions. Also, these models process the features of the encoder and decoder separately, which promotes cross-necessary information loss. The processed information of encoders and decoders is complementary to each other and they need to be managed together. \\
\begin{figure}[t]
\begin{center}
\includegraphics[width=0.99\linewidth]{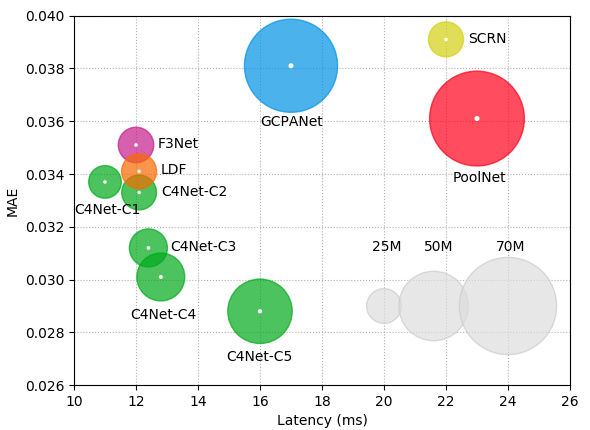}
\end{center}
\caption{Performance comparison between our proposed and other state-of-the-art methods by their complexity, latency (ms) and MAE results on DUTS-Test dataset.}
\label{fig:complexities}
\end{figure}
By analyzing the visual results of different models, we have noticed there are some incorrect predictions like random holes or excessive parts. These types of issues often come from the lack of high-level semantic information. After a couple of comprehensive experiments, we have come up with a modification of the pyramid-pooling module~\cite{PoolNet} and called it a pyramid-semantic module, which contains multi-scale context-aware feature representation and channel-wise shift and attention.\\
We proposed a complementary extraction module to combine and process low and high-level information by taking into account the feature representations of the edges. Contextual compression module is made to compress the connections between encoder and decoder, to maintain necessary features, and make the processing faster. Also, our proposed loss function helps to minimize false predictions and is formulated by false-positive values for each pixel. So our contributions are the following.

\begin{itemize}
  \item Proposed a family of \textit{C$^{4}$Net} architectures with different complexities and performance.
  \item Proposed the \textit{Complementary Extraction Module} to combine and process low and high-level information with edge preservation.
  \item Proposed \textit{Excessiveness Loss} function to minimize false-positive predictions.
\end{itemize}

\section{Related Work}
During recent decades, researchers developed algorithms for the salient object detection problem, which are based on hand-crafted features (\eg color or texture) \cite{old_sod1, old_sod2, old_sod3, old_sod4}. Most of these traditional methods have been surpassed by convolutional neural networks (CNN) in terms of quality and speed. Deep convolutional networks are capable to extract necessary semantic information and combine them properly. By bringing all known approaches together, we can formulate the following common factors. \\
\textbf{Features Processing Methods}. As the majority of the recent solutions are based on U-Net-like architectures, one of the most important parts is the feature processing part at every layer of the decoder. Zuyao Chen \etal \cite{GCPANet} considered separate processing. They applied multiplication operation between the encoder's and decoder's features, then concatenated the results of different branches. Jun Wei \etal \cite{F3Net} also proposed almost the same approach, where their model separately processes encoder's and decoder's features, multiplies them as a fusing operation then they apply an addition operation on it. They considered that the encoder's features are low-level representations and contain detailed information. In contrast, the decoder's features are high-level representations and contain rich semantic information. By multiplying them, they clean the information and add complementary features for high-quality detection. We find these approaches have some drawbacks, because of the separate processing of high and low-level features, which causes cross-necessary information loss. \\
\textbf{Edge Preservation Approaches}. The salient object detection problem requires a solution with high-quality detection, especially on edges, which need to be purified and exquisite. To maintain the high quality of edges, Jiang-Jiang Liu \etal \cite{PoolNet} apply additional edge detection on the middle layers of the decoder. Mengyang Feng \etal \cite{AFNet} proposed the \textit{Boundary-Enhanced Loss} function for shallower layers of the network, which is responsible for extracting the features and purifying the edges. All these methods tend to improve the quality of edges by using additional and separated edge-specific processing. We find that the core loss functions also must be adjusted for this manner and proposed weighted losses, where edge pixels have high weights. A similar solution has been done in methods like ~\cite{F3Net}. \\
\textbf{Global Features Extraction}. As we have mentioned in the previous sections, one of the advantages of deep learning solutions is the capability of semantic information extraction, but the majority of these approaches do not have direct ties between global semantic modules and shallower layers. Jiang-Jiang Liu \etal \cite{PoolNet} and Zuyao Chen \etal \cite{GCPANet} gave solutions for this manner. They designed a module for high-level semantic information and a global guiding flow to distribute the information in top layers. Our contribution tends to improve global information extraction by channel-wise attention and shifting.

\begin{figure*}[t]
\begin{center}
\includegraphics[width=0.999\linewidth]{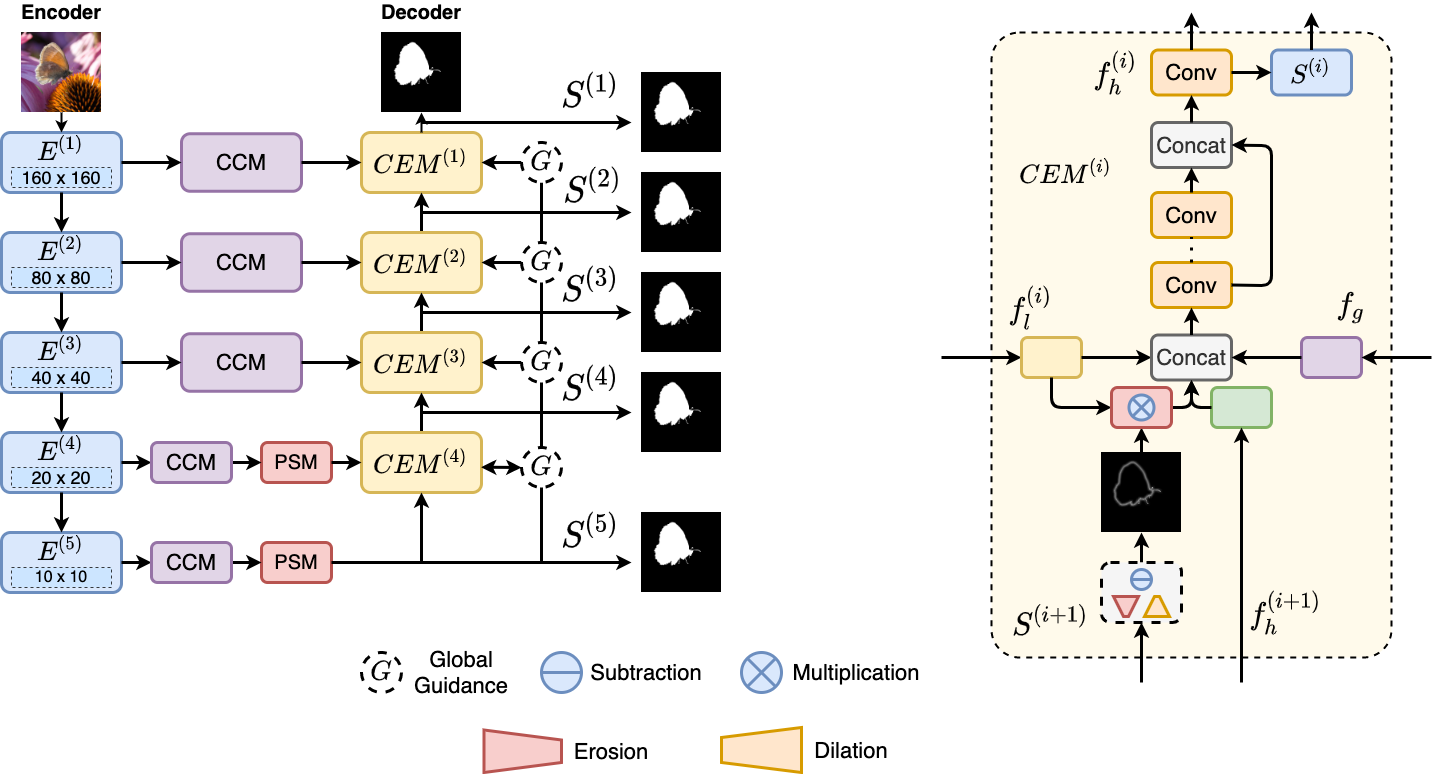}
\end{center}
   \caption{An overview of our proposed C$^{4}$Net architecture. The model is based on a \textit{ResNet} \cite{ResNet} encoder with multilevel supervision $S^{(i)}$. Contextual Compression Module (CCM) is used as compressed shortcut connection between encoder and decoder. Pyramid-Semantic Module (PSM) is used to extract high level semantic information, which is also used in Global Guidance flow (G) and Complementary Extraction Module (CEM) is used to combine three feature representations from the encoder, decoder and guiding flow.}
\label{fig:network}
\end{figure*}

\section{Proposed Method}
\label{pm}
In this section, we will describe the architecture of our model and all consisting parts of it. Before going into details, we first need to refer to the feature combination approach. As U-Net-like architectures have shortcut connections between encoder and decoder, it is very important how their feature representations will be combined. Based on the conducted experiments and the design choice of our model, we found out that joint processing of encoder's and decoder's features works better than the branched-separated approach. Also, the concatenation aggregation function for skip connections and feature combinations leads better results compared with other aggregation functions like addition and multiplication. The details about those approaches can be found in the ablation study (Section ~\ref{abl}).

\subsection{Contextual Compression Module}
As we noted in the previous section, shortcut connections between encoder and decoder are crucial for high-quality detection and exquisite boundaries. We defined a new term called Compression Factor (CF) to compress half of channels of the network, which starts from the shortcut connections.
\begin{equation}
    f_{l} = \delta(BN(Conv_{cf}(f^{e}_{l})))
\end{equation}
where $f^{e}_{l}$ is the feature representation of the encoder, $Conv_{cf}$ is a convolution layer with $cf$ filters (\{32, 64, 128\} in our experiments), $BN$ is a batch normalization layer, $\delta$ is the ReLU activation function. By conducting a couple of comprehensive experiments for shortcut connections, we came up with the following conclusion.

\begin{itemize}
  \item Without any additional compression block, the encoder provides noisy information. The experimental results are reported in the fifth and last row of Table~\ref{tab:modules_effectivness}.
  \item By using feature compression for half of the network, we have increased the speed of training and testing regimes, also, optimized the memory allocation of it.
\end{itemize}

We use the same compression factor for all layers of the CCM module and for the decoder of our network. Table~\ref{tab:model_architectures} contains an ablation study for different compression factors and Table~\ref{tab:modules_effectivness} shows the effectiveness of the CCM module. Each next row of Table~\ref{tab:modules_effectivness} either is built on top of the previous row or is replaced with the mentioned module. 

\begin{table}
\begin{center}
\begin{tabular}{|c|c|c|c|c|c|}
\hline
Name & Encoder & CF & $MAE$ & $mF$ & $E_{\xi}$ \\
\hline\hline\rule{0pt}{2.2ex}
\textbf{C$^{4}$Net-C1} & R34 & 64 & 0.033 & 0.872 & 0.925 \\
\textbf{C$^{4}$Net-C2} & R50 & 32 & 0.033 & 0.867 & 0.928 \\
\textbf{C$^{4}$Net-C3} & R50 & 64 & 0.031 & 0.872 & 0.929 \\
\textbf{C$^{4}$Net-C4} & R50 & 128 & 0.030 & 0.875 & 0.929 \\
\textbf{C$^{4}$Net-C5} & R101 & 64 & 0.029 & 0.886 & 0.934 \\
\hline
\end{tabular}
\end{center}
\caption{Architectures of our proposed model with different complexities. Results are based on \textit{DUTS-Test} dataset.}
\label{tab:model_architectures}
\end{table}

\subsection{Pyramid-Semantic Module}
\begin{figure}[t]
\begin{center}
\includegraphics[width=0.98\linewidth]{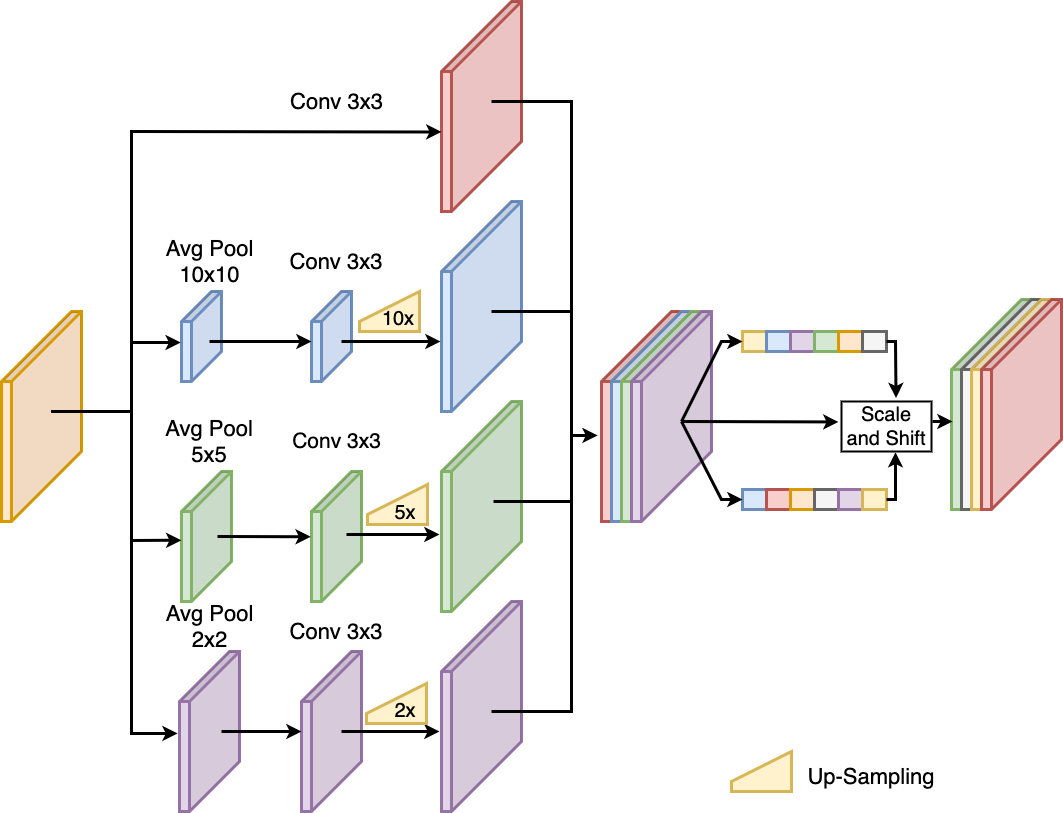}
\end{center}
    \caption{Illustration of our Pyramid-Semantic Module.}
\label{fig:psm}
\end{figure}

We have noted the importance of low-level information, but high-level information is also very important for better detection. We have modified the PPM module proposed in~\cite{PoolNet} and designed the Pyramid-Semantic module, which is responsible to handle and maintain high-level information. PSM consists of two main parts, which are visualized in Figure~\ref{fig:psm}. \\
The first part is responsible for pyramid-feature extraction, which is made by branched pooling operation and the second part is a channel-wise attention module, which scales and shifts feature representations by channels.
For the first part of this module, we use average pooling operation to get multi-resolution pyramid features. Also, It has an identical branch, which lets to maintain high-level semantic information of the same resolution.
\begin{equation}
    \bar{f}^{(j)}_{h} = Up(Conv(Pool_{k}(f_{l})))
\end{equation}
where $f_{l}$ is the input feature representation of deeper layers provided by the CCM module, $Pool_{k}$ is the average pooling operation with kernel size $k$. For each branch $(j\in\{2, 3, 4\})$ we use different kernel sizes: \textit{\{10, 5, 2\}}. $Conv$ is a combination of convolution, batch normalization, and a ReLU activation functions. $Up$ is a \textit{bilinear} up-sampling operation.

At the end of the first part, all four branches get concatenated into one feature representation.\\
The second part is the channel-wise attention module, which scales and shifts features. Let $f_{h}\in R^{H\times W\times C}$ be the input features. We apply global pooling operation on $f_{h}$ and get $\tilde{f} \in R^{C}$.

\begin{equation}
\begin{split}
    w &= \sigma(fc_{2}(\delta(fc_{1}(\tilde{f}, W^{(w)}_{1})), W^{(w)}_{2})) \\
    v &= \sigma(fc_{2}(\delta(fc_{1}(\tilde{f}, W^{(v)}_{1})), W^{(v)}_{2}))
\end{split}
\end{equation}

\begin{equation}
    f_{h} = w \odot f_{h} \oplus v, \;\;\;\;w,v\in{R^{C}}
\end{equation}
where $fc_{1}$ and $fc_{2}$ are fully connected layers, $W^{(w)}$ and $W^{(v)}$ are weights matrices for scaling and shifting respectively. $\delta$ is the ReLU activation function, and $\sigma$ is the Sigmoid activation function, $\odot$ is channel-wise multiplication and $\oplus$ is channel-wise addition function. \\
To verify the effectiveness of our modification, we have conducted two other experiments as well with other similar solutions like \textit{Pyramid-Pooling Module (PPM)}~\cite{PoolNet} and \textit{Atrous Spatial Pyramid Pooling (ASPP)}~\cite{DeepLab_ASPP}. Table~\ref{tab:modules_effectivness} shows that our method outperforms other pyramid-based approaches mainly because of the channel-wise shift and attention.

\begin{table}
\begin{center}
\begin{tabular}{|c|c|c|c|c|}
\hline
Modules & $MAE$ & $mF$ & $E_{\xi}$ \\
\hline\hline
Baseline (w/o shortcut) & 0.045 & 0.804 & 0.903 \\
EL & 0.043 & 0.813 & 0.904 \\
EL + CCM & 0.034 & 0.851 & 0.917 \\
EL + CCM + CEM & 0.032 & 0.867 & 0.921 \\
EL + CCM + CEM + PSM & 0.031 & 0.872 & 0.929 \\
\hline\rule{0pt}{2.2ex}
EL + CCM + CEM + PPM & 0.033 & 0.865 & 0.923 \\
EL + CCM + CEM + ASPP & 0.033 & 0.867 & 0.924 \\
\hline\rule{0pt}{2.2ex}
CCM + CEM + PSM & 0.033 & 0.867 & 0.922 \\
EL + CEM + PSM$^{\dag}$ & 0.033 & 0.874 & 0.924 \\
\hline
\end{tabular}
\end{center}
\caption{Ablation study for different modules of C$^{4}$Net-C3 architecture. Symbol $\dag$ indicates a model with straight shortcut connections w/o any operation. The results are based on \textit{DUTS-Test} dataset.}
\label{tab:modules_effectivness}
\end{table}

\subsection{Complementary Extraction Module}
As we have referred to the structure of feature processing approach at the beginning of Section~\ref{pm}, it is crucial to choose a right extraction mechanism and a combination function, thus we have decided to choose the joint processing method (PipeMode) with concatenation combination function, as they perform better for the model design we have chosen (see Section~\ref{abl}). As a summary, we have designed a module, which is responsible for three different feature extraction. $f_{l} \in R^{H\times W\times C}$ is the low-level feature representation, which contains rich details with noise, in contrast to $f_{h} \in R^{H\times W\times C}$, which is a high-level feature and does not contain rich details for exquisite detection, but it is noisy-free. $f_{g} \in R^{H\times W\times C}$, where $f_{g} = f^{4}_{h} + f^{5}_{h}$ is a global guidance flow using the fourth and fifth layer's features, which helps to complement high-level semantic information in shallow layers. \\
As we seek high-quality detection, we also need to reduce the error on edges, because edges contain the most errors of the detection as shown in \cite{LDF}. To adjust the model for multi-level supervision, we apply a convolution layer with the sigmoid function on the output of the previous layer and get $S^{(i)}$ binary mask.
We compute edges by dilating and eroding the $S^{(i)}$ mask, then apply pixel-wise multiplication with $f_{l}$. $f_{edge}$ will be the edge feature representation. We concatenate all four features, $f_{l}, f_{h}, f_{edge}$ and $f_{g}$ and feed into a convolution block with six convolutions, batch normalization, ReLU layers, and a skip connection. Figure~\ref{fig:network} contains the visualization of our proposed CEM module and Table~\ref{tab:modules_effectivness} shows its effectiveness with three different metrics.

\begin{table*}
\setlength{\tabcolsep}{4.6pt}
\small
\begin{center}
\begin{tabular}{|r|c|c|c|c|c|c|c|c|c|c|c|c|c|c|c|}
\hline
\multirow{3}{5em}{\textbf{Algorithms}}
& \multicolumn{3}{|c}{\textbf{ECSSD}} 
& \multicolumn{3}{|c}{\textbf{HKU-IS}} 
& \multicolumn{3}{|c}{\textbf{PASCAL-S}} 
& \multicolumn{3}{|c}{\textbf{DUT-OMRON}} 
& \multicolumn{3}{|c|}{\textbf{DUTS-Test}} \\
& \multicolumn{3}{|c}{1,000 images} 
& \multicolumn{3}{|c}{4,447 images} 
& \multicolumn{3}{|c}{850 images} 
& \multicolumn{3}{|c}{5,168 images} 
& \multicolumn{3}{|c|}{5,019 images} \\
\cline{2-16}
 & MAE & $mF$ & $E_{\xi}$ 
& MAE & $mF$ & $E_{\xi}$
& MAE & $mF$ & $E_{\xi}$
& MAE & $mF$ & $E_{\xi}$
& MAE & $mF$ & $E_{\xi}$ \\
\hline\hline\rule{0pt}{2.2ex}
PiCANet (CVPR2018) & .046 & .919 & .951 & .043 & .900 & .947 & .075 & .831 & .893 & .065 & .759 & .860 & .050 & .828 & .909 \\

DGRL (CVPR2018) & .045 & .910 & .945 & .037 & .897 & .947 & .074 & .819 & .882 & .063 & .738 & .848 & .051 & .802 & .892 \\

EGNet (ICCV2019) & .041 & .933 & .953 & .031 & .917 & \textcolor{mblue}{\textbf{.956}} & .074 & .833 & .885 & .053 & .767 & .857 & .039 & .856 & .915 \\

SCRN (ICCV2019) & .037 & \textcolor{mblue}{\textbf{.935}} & .954 & .034 & .917 & .953 & .063 & \textcolor{mblue}{\textbf{.850}} & \textcolor{mblue}{\textbf{.902}} & .056 & .772 & \textcolor{mblue}{\textbf{.863}} & .039 & \textcolor{mblue}{\textbf{.860}} & .915 \\

CPD (CVPR2019) & .037 & .923 & .950 & .034 & .905 & .948 & .070 & .828 & .884 & .056 & .754 & .850 & .043 & .836 & .906 \\

BASNet (CVPR2019) & .037 & .927 & .950 & .032 & .914 & .950 & .076 & .824 & .879 & .056 & \textcolor{mblue}{\textbf{.773}} & \textcolor{mgreen}{\textbf{.864}} & .047 & .832 & .897 \\

PoolNet (CVPR2019) & .038 & .931 & .955 & .030 & .917 & .955 & .065 & .846 & .900 & .054 & .756 & .849 & .036 & .854 & .917 \\

MINet (CVPR2020) & .034 & .931 & \textcolor{mblue}{\textbf{.956}} & .029 & \textcolor{mblue}{\textbf{.918}} & \textcolor{mgreen}{\textbf{.959}} & .064 & .836 & .896 & .056 & .764 & .861 & .037 & .854 & \textcolor{mblue}{\textbf{.920}} \\

LDF (CVPR2020) & .034 & .930 & .926 & \textcolor{mgreen}{\textbf{.027}} & .914 & .954 & \textcolor{mblue}{\textbf{.060}} & .848 & .866 & \textcolor{mblue}{\textbf{.052}} & \textcolor{mblue}{\textbf{.773}} & .862 & \textcolor{mblue}{\textbf{.034}} & .855 & .910 \\

GCPANet (AAAI2020) & .035 & .929 & .954 & .031 & .917 & \textcolor{mblue}{\textbf{.956}} & .063 & .840 & .898 & .057 & .767 & .857 & .038 & .858 & .919 \\

F$^{3}$Net (AAAI2020) & \textcolor{mblue}{\textbf{.033}} & .925 & .930 & \textcolor{mblue}{\textbf{.028}} & .912 & .954 & .062 & .834 & .886 & .053 & .770 & .862 & .035 & .842 & .903 \\

\textbf{C$^{4}$Net-C4 (Ours)} & \textcolor{mgreen}{\textbf{.029}} & \textcolor{mgreen}{\textbf{.939}} & \textcolor{mgreen}{\textbf{.957}} & \textcolor{mgreen}{\textbf{.027}} & \textcolor{mgreen}{\textbf{.925}} & \textcolor{mblue}{\textbf{.956}} & \textcolor{mgreen}{\textbf{.056}} & \textcolor{mgreen}{\textbf{.854}} & \textcolor{mgreen}{\textbf{.903}} & \textcolor{mgreen}{\textbf{.051}} & \textcolor{mgreen}{\textbf{.776}} & 
\textcolor{mblue}{\textbf{.863}} & \textcolor{mgreen}{\textbf{.030}} & \textcolor{mgreen}{\textbf{.875}} & \textcolor{mgreen}{\textbf{.929}} \\

\hline\rule{0pt}{2.2ex}
\textbf{C$^{4}$Net-C5 (Ours)} & .030 & \textcolor{mred}{\textbf{.939}} & .956 & \textcolor{mred}{\textbf{.025}} & \textcolor{mred}{\textbf{.931}} & \textcolor{mred}{\textbf{.961}} & \textcolor{mred}{\textbf{.055}} & \textcolor{mred}{\textbf{.861}} & \textcolor{mred}{\textbf{.904}} & \textcolor{mred}{\textbf{.047}} & \textcolor{mred}{\textbf{.788}} & \textcolor{mred}{\textbf{.865}} & \textcolor{mred}{\textbf{.029}} & \textcolor{mred}{\textbf{.886}} & \textcolor{mred}{\textbf{.937}} \\

\hline
\end{tabular}
\end{center}
\caption{Performance comparison with 11 state-of-the-art methods on 5 benchmark datasets. MAE (smaller is better), mean F-measure ($mF$ larger is better), E-measure ($E_{\xi}$ larger is better) metrics used to evaluate the results. All models are based on ResNet backbones. The best and the second best results among models with ResNet50 backbones highlighted in \textcolor{mgreen}{\textbf{green}} and \textcolor{mblue}{\textbf{blue}} respectively. The \textcolor{mred}{\textbf{red}} color indicates the best results of our model with ResNet101 backbone.}
\label{tab:comparison}
\end{table*}

\begin{figure*}[t]
\begin{center}
\includegraphics[width=0.999\linewidth]{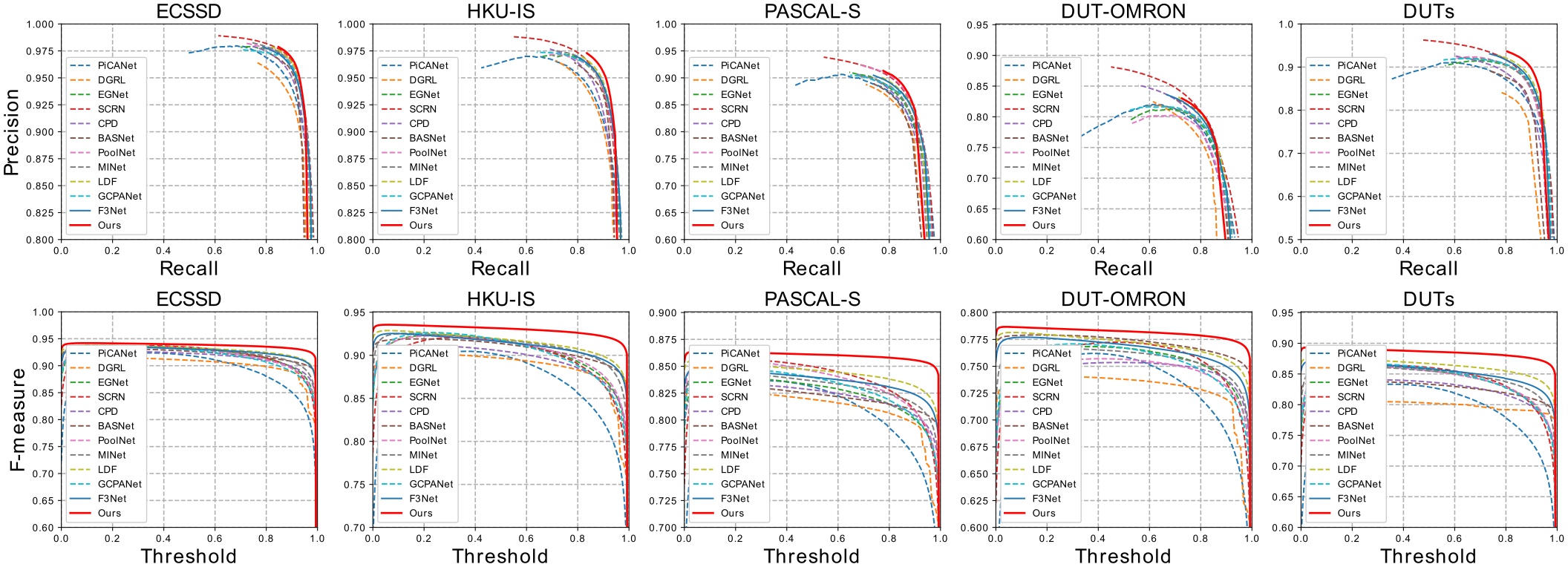}
\end{center}
   \caption{Illustration of PR curves (the first row), F-measure curves (the second row) on 5 benchmark datasets.}
\label{fig:pr_curve}
\end{figure*}

\subsection{Excessiveness Loss Function}
The binary cross-entropy function is one of the most widely used loss functions in salient object detection problem, however, the equal treatment of pixels and the ignorance of global structures make the function not effective, so we have used the approach proposed by Jun Wei~\etal\cite{F3Net}. \\
First, we have used weighted losses by adding pixel-wise weight matrix to increase the importance of the edges.

\begin{equation}
    \omega_{i,j} = 1 + \tilde{\lambda}\; \Big|\dfrac{1}{N}\sum^{k,k}_{\tilde{i}=1, \tilde{j}=1} Gt_{\tilde{i},\tilde{j}} - Gt_{i,j}\Big|
\end{equation}
where $Gt$ is the ground truth mask, $\tilde{\lambda}$ is a hyper parameter, $|.|$ is the absolute value, $N=k\times{k}$ is the number of pixels of the window with kernel size $k$.

\begin{equation}
\fontdimen14\textfont2=6pt
\fontdimen17\textfont2=4pt
\textstyle
    L_{bce}^{i,j} = Gt_{i,j}*\log{(S_{i,j})} + (1 - Gt_{i,j})*\log{(1 - S_{i,j})}
\end{equation}
\begin{equation}
    L_{wbce} = -\;\dfrac{\sum_{i=1,j=1}^{W,H}\;\omega_{i,j}*L_{bce}^{i,j}}{\sum_{i=1,j=1}^{W,H}\;\omega_{i,j}}
\end{equation}
where $S_{ij}$ is the prediction of $i,j$-th pixel and $W,H$ is the width and height of the output. \\
As we seek a solution with structural preservation of the objects, we use the following loss.  
\begin{equation}
    L_{wiou} = 1 - \dfrac{\sum^{W,H}_{i=1,j = 1} S_{i,j}*Gt_{i,j}*\omega_{i,j}}{\sum^{W,H}_{i=1,j=1} (S_{i,j} + Gt_{i,j} - S_{i,j}*Gt_{i,j})*\omega_{i,j}}
\end{equation}
$L_{wiou}$ is the weighted intersection over union loss function, which handles global structures of the foreground object and increases the impact of the pixels near the edges.\\
To improve the detection results, we proposed a new loss function, which is called the  \textit{Excessiveness Loss} function. We observe \textit{false positive} ($FP$), \textit{false negative} ($FN$), and \textit{true positive} ($TP$) predictions for each example. By analyzing the error for these values, we found out that \textit{the error mostly accumulates by excessive predictions}, which means in most cases,
$FP$ is higher than $FN$. Table~\ref{tab:el_contrib} contains an ablation study, where $mFP$ and $mFN$ are normalized by the resolution of the output mask. Our proposed weighted \textit{excessiveness loss} function is
\begin{equation}
\begin{split}
    \omega{TP} = \sum^{W, H}_{i=1, j=1}S_{i,j}*Gt_{i,j}*\omega_{i,j} \\
    \omega{FP} = \sum^{W, H}_{i=1, j=1}\delta{(S_{i, j}- Gt_{i, j}) * \omega_{i, j}}
\end{split}
\end{equation}
\begin{equation}
    L_{wel} = \dfrac{\omega{FP}}{\omega{FP} + \gamma \omega{TP}}
\end{equation}
where $\omega{FP}, \omega{TP}$ are weighted \textit{false positive} and \textit{true positive} respectively and they are differentiable, $\delta$ is the ReLU activation function, $\gamma$ is a hyper-parameter.  \\
The overall loss of our model is a weighted combination of these three loss functions.
\begin{equation}
    L = L^{(1)}_{wel} + \sum^{5}_{i=1}\dfrac{1}{2^{i-1}} (L^{(i)}_{wbce}+L^{(i)}_{wiou})
\end{equation}
where $L^{(i)}_{wel}, L^{(i)}_{wbce}$ and $L^{(i)}_{wiou}$ are the corresponding loss functions for $i$-th layer. To verify the effectiveness of $EL$ loss function, we show the results of two-way experiments in the first and last group of Table~\ref{tab:modules_effectivness}.

\section{Experiments}
\subsection{Datasets and Evaluation Metrics}
\label{seq:dataset_metrics}
To evaluate our proposed method, we have used five popular benchmark datasets, including ECSSD \cite{ECSSD} with 1000 images, PASCAL-S \cite{PASCAL-S} with 850 images, HKU-IS \cite{HKU-IS} with 4447 images, DUT-OMRON \cite{DUT-OMRON} with 5168 images, and DUTS \cite{DUTs} with 15572 images. DUTS is currently the biggest  salient object detection dataset and contains 10553 training and 5019 testing examples. We used only the DUTS-Train dataset for training and others for testing. Three metrics are used to evaluate the performance of our model and other state-of-the-art methods. The first metric is the \textit{Mean Absolute Error (MAE)}, which is one of the most commonly used metrics for salient object detection, as shown in Eq.~\ref{eq:mae}. Another widely used metric of the SOD problem is the \textit{mean F-measure} $(mF)$ which is formulated with \textit{precision} and \textit{recall} as the $F_{\beta}$ score, where $\beta$ = 0.3. We also use the E-measure $(E_{\xi})$ \cite{Emeasure} metric, which uses the combination of local pixel values and their means to evaluate the similarity between prediction and ground truth masks.

\begin{table}[t]
\footnotesize
\begin{center}
\begin{tabular}{|c|c|c|c|c|}
\hline
Modules & $MAE$ & $mF$ & $mFP$ & $mFN$ \\
\hline\hline
CCM + CEM + PSM & 0.033 & 0.867 & 0.022 & 0.016 \\
EL + CCM + CEM + PSM & 0.031 & 0.872 & 0.019 & 0.016 \\
\hline
\end{tabular}
\end{center}
\caption{Ablation study of $EL$ contribution of C$^{4}$Net-C3 architecture on \textit{DUTS-Test} dataset.}
\label{tab:el_contrib}
\end{table}

\begin{equation}
\label{eq:mae}
    MAE = \dfrac{1}{W\times{H}} \sum^{W}_{i=1}\sum^{H}_{j=1} |P_{i, j} - Gt_{i, j}|
\end{equation}
where $P$ is the prediction and $Gt$ is the ground truth mask. To show the robustness of our proposed method, we also plot the \textit{Precision-Recall (PR)} curve, which is calculated by sliding the threshold from 0 to 1. The larger the area under the PR curve, the better is the performance.

\subsection{Implementation Details}
We use DUTS-Train as the training dataset, with randomly cropping and horizontal flipping augmentation techniques. Different architectures of ResNet \cite{ResNet} pre-trained on \textit{ImageNet} are used as the encoder and other parts of the model are initialized randomly from the uniform distribution. We use an adaptive learning rate with a maximum value of 0.005 for the encoder and 0.05 for other parts. The network is trained with \textit{stochastic gradient descent (SGD)} with 0.9 momentum and 0.0005 weight decay parameter values. The batch size is set to 20 with 50 epochs. Our network is implemented with Pytorch v1.6 and the training and testing processes are conducted on an \textit{Nvidia RTX 2080 ti} GPU and \textit{Intel Core i9-9900k} CPU
device. The performance report of our proposed models can be found in Figure~\ref{fig:complexities}, where our fastest model has about 11ms (90 fps) inference time by surpassing other state-of-the-art solutions. All images are resized to 320$\times$320 during training and testing, without any post-processing approach.

\begin{figure*}[t]
\begin{center}
\includegraphics[width=0.99\linewidth]{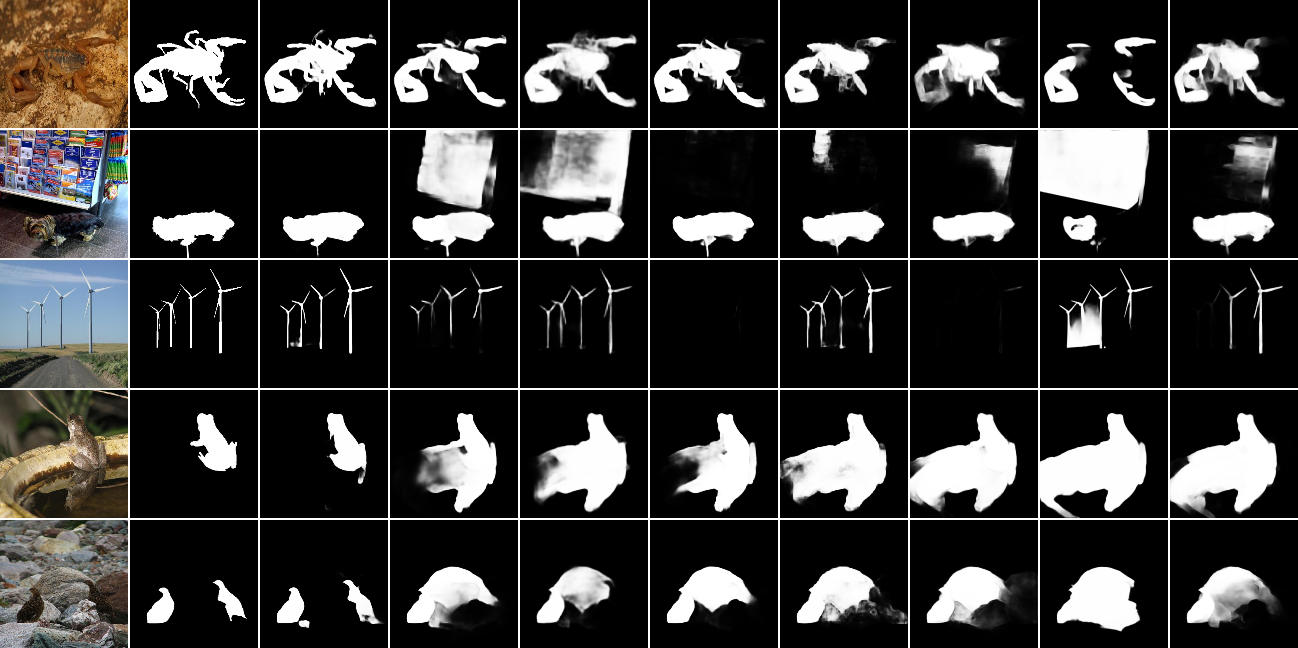}
\end{center}
\vspace{-10pt}
\textbf{\hspace{0.4cm} Image \hspace{0.8cm} GT \hspace{1cm} Ours \hspace{0.7cm} F$^{3}$Net \hspace{0.3cm} GCPANet \hspace{0.5cm} LDF \hspace{0.7cm} MINet \hspace{0.5cm} PoolNet \hspace{0.4cm} BASNet \hspace{0.4cm} EGNet}
\vspace{6pt}
   \caption{Qualitative comparison of the proposed model with other state-of-the-art methods. Our model has minimal false prediction and the edges are more exquisite than others.}
\label{fig:vis_comparison}
\end{figure*}

\subsection{Comparison with State-of-the-Arts}
We compare our proposed algorithm with 11 state-of-the-art methods, including PiCANet-R \cite{PiCANet}, DGRL \cite{DGRL}, EGNet \cite{EGNet}, SCRN \cite{SCRN}, CPD \cite{CPD}, BASNet \cite{BASNet}, PoolNet \cite{PoolNet}, MINet \cite{MINet-CVPR2020}, LDF \cite{LDF}, GCPANet \cite{GCPANet} and F$^{3}$Net \cite{F3Net}, which all have ResNet backbones. For fair comparison, we use the same three metrics for all methods with the same script and visualize their PR and $F_{\beta}$ curves. \\
\textbf{Quantitative Comparison}. To compare all listed methods with the metrics mentioned in Section~\ref{seq:dataset_metrics}, we have created Table~\ref{tab:comparison} with 11 state-of-the-art algorithms. Without bells and whistles, our proposed method works better on all benchmark datasets. One of the highest improvements among models with ResNet50~\cite{ResNet} backbone is on the DUTS-Test dataset, where our algorithm boosts the results by more than 11\% in terms of the MAE. Our \textit{C$^{4}$Net-C1} model, which has ResNet34~\cite{ResNet} as a backbone, also surpasses many SOTA results. There is a comparison between our proposed architectures and other solutions by their latency (ms), MAE on DUTS-Test, and number of parameters in Figure~\ref{fig:complexities}. Some other SOTA models like MINet~\cite{MINet-CVPR2020} or BASNet~\cite{BASNet} are either too slow or too complex and are out of the chosen window. Also, the PR and $F_{\beta}$ curves show the robustness of our models on the mentioned datasets. \\
\textbf{Qualitative Comparison}. The visual comparison examples are shown in Figure~\ref{fig:vis_comparison}, where our model has better quality on the edges among all algorithms, because of our weighted loss functions for all layers and the complimentary edge-feature extraction in our CEM module, which helps to get exquisite boundaries. Our model is able to detect narrow parts and recover lost information. Another visible improvement is minimal false predictions, especially false-positive values. Our proposed EL function is able to minimize false-positive areas and maximize the true positive predictions, which is visible in the comparison figure.

\subsection{Ablation Study}
\label{abl}
\textbf{Features Combination Methods}. Modern architectures of deep learning solutions for the SOD problem are based on encoders, decoders, and shortcut connections between them. Each layer of the decoder is responsible for combining features of decoder and encoder. There are two main approaches to using that information. The first one is by simply concatenating them, the second one is to fuse them by using other functions like multiplication or addition. Another very important thing is  the features processing structure at every layer of the decoder. Methods like \cite{GCPANet, F3Net} prefer to process encoder's and decoder's features separately and then fuse them by multiplication or addition. \\
We tried to find out the answers to two main questions
\begin{itemize}
  \item How the features of the encoder and decoder need to be processed. (joint or separated)
  \item How they need to be combined. (fusing or concatenating)
\end{itemize}

\begin{figure}[t]
\begin{center}
\includegraphics[width=0.9\linewidth]{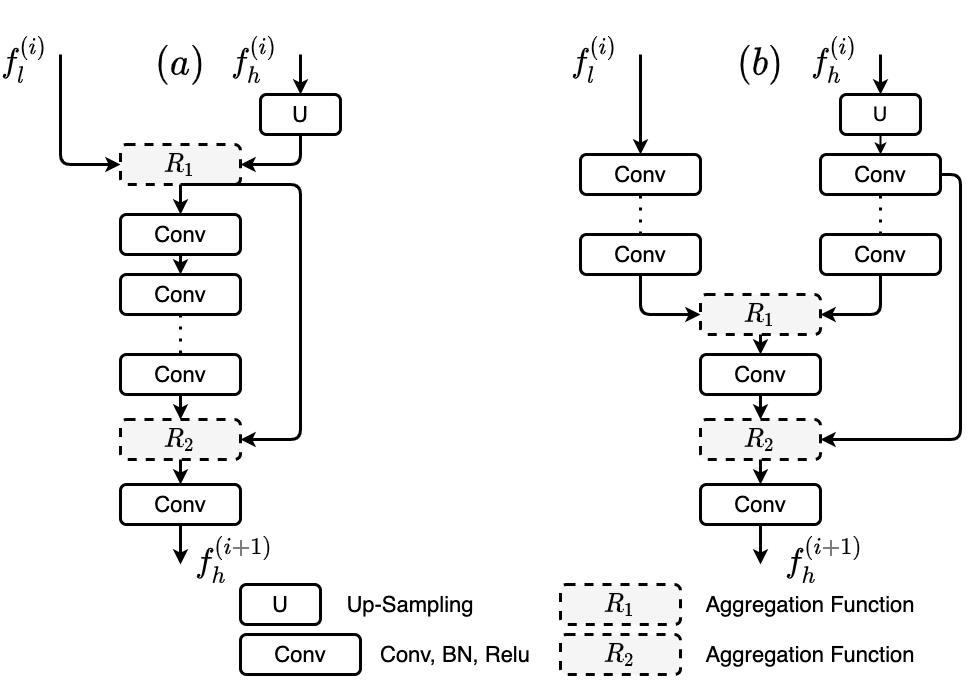}
\end{center}
   \caption{An overview of our proposed structures for a layer of a decoder. $(a)$ is the joint-features module and we called it PipeMode and $(b)$ is the separated-features module and we called it BranchedMode. $R_{1}$ and $R_{2}$ are aggregation functions.}
\label{fig:pipe_branch}
\end{figure}

To answer these questions, we proposed two main structures of a decoder's layer Figure ~\ref{fig:pipe_branch}. We designed a PipeMode layer, which is a joint processing of two feature representations, and BranchedMode layer, which is separated processing of the features. They both contain two aggregation functions $R_{1}$ and $R_{2}$.
\\The features of encoding layers, $f^{(i)}_{l}$ contain low-level information like edges, tiny areas, and high-frequency data, which is crucial for high-quality detection, especially on edges and they contain noisy information. The feature representations of decoding layers, $f^{(i)}_{h}$ contain noisy free high-level information like class, position, or shape of the object. These features representations helped Jun Wei~\etal \cite{F3Net} to propose a solution like our BranchedMode, where they used the multiplication function to fuse high and low-level features and to clean noisy parts, then they added a skip-like connection as complementary information. We find this approach has some drawbacks, because of the choice of the structure and aggregating functions.

To find out the real behavior of these two approaches, we made different experiments with these structures by using different aggregating functions. We designed a simple architecture with \textit{ResNet50} backbone, our proposed structures for the decoder, and shortcut connections between them. The BCE loss function was used on top of the decoder. Each model was trained three times with randomly initialized weights and the result was calculated by taking the average of the best performance at each run. Table~\ref{tab:pipe_branch} contains reports of our experiments.
Based on those results and the architecture choice, we can say:
\begin{itemize}
  \item In general, PipeMode gives better results than BranchedMode, which shows that the sharing of information about different features leads to better results.
  \item The concatenation aggregation function works better for both structures.
\end{itemize}
PipeMM results are missing in Table~\ref{tab:pipe_branch}, because it disturbs the training of the model with activation values pushed to zero~\cite{glorot}.

\begin{table}[t]
\begin{center}
\begin{tabular}{|c|c|c|c|c|c|}
\hline
Name & $R_{1}$ & $R_{2}$ & $MAE$ & $mF$ & $E_{\xi}$ \\
\hline\hline\rule{0pt}{2.2ex}
BranchPP & Plus & Plus & 0.0356 & 0.8514 & 0.9162 \\

BranchMM & Mul & Mul & \textcolor{mred}{\textbf{0.0358}} & \textcolor{mred}{\textbf{0.8496}} & 0.9165 \\

BranchCC & Cat & Cat & \textcolor{mblue}{\textbf{0.0353}} & \textcolor{mgreen}{\textbf{0.8535}} & \textcolor{mblue}{\textbf{0.9168}} \\

BranchMP & Mul & Plus & \textcolor{mblue}{\textbf{0.0353}} & 0.8509 & \textcolor{mred}{\textbf{0.9152}} \\

PipePP     & Plus & Plus & 0.0348 & \textcolor{morange}{\textbf{0.8528}} & 0.9164\\

PipeMM     & Mul & Mul & - & - & - \\

PipeCC     & Cat & Cat & \textcolor{mgreen}{\textbf{0.0342}} & 0.8512 & \textcolor{mgreen}{\textbf{0.9171}} \\

PipeCP     & Cat & Plus & 0.0347 & 0.8520 & 0.9159 \\
\hline
\end{tabular}
\end{center}
\caption{Results of our proposed structures on \textit{DUTS-Test} dataset with different aggregating functions, where Plus is addition, Mul is multiplication and Cat is concatenation. The \textcolor{mgreen}{\textbf{green}} is the overall best result, \textcolor{mred}{\textbf{red}} is the overall worst result,
\textcolor{mblue}{\textbf{blue}} is the best result among BranchedModes,
\textcolor{morange}{\textbf{orange}} is the best result among PipeModes.}
\label{tab:pipe_branch}
\end{table}

\section{Conclusion}
In this paper, we proposed a new solution for the salient object detection problem. Our ablation study contains an investigation about feature combination approaches, where we showed that joint learning with pipe-mode works better than branched-mode. Also, based on our model design, the feature concatenation gives better results for both structures. Our proposed solution is able to extract and preserve feature representations on edges (\textit{Complementary Extraction Module, CEM}) with high-level semantic information (\textit{Pyramid-Semantic Module, PSM}), which leads to high-quality detection. Also, the proposed weighted \textit{Excessiveness Loss (EL)} function helps to minimize false prediction values. Each module leads to significant improvement, which is shown in Table~\ref{tab:modules_effectivness}. The comparison with 11 state-of-the-art methods shows that our approach outperforms other solutions on all benchmark datasets under three evaluation metrics.

{\small
\bibliographystyle{ieee_fullname}
\bibliography{main}

\begin{thebibliography}{10}\itemsep=-1pt

\bibitem{DeepLab_ASPP}
Liang-Chieh Chen, George Papandreou, Iasonas Kokkinos, Kevin Murphy, and Alan
  L~Yuille.
\newblock Deeplab: Semantic image segmentation with deep convolutional nets,
  atrous convolution, and fully connected crfs.
\newblock In {\em IEEE transactions on pattern analysis and machine
  intelligence}, 2018.

\bibitem{GCPANet}
Zuyao Chen, Qianqian Xu, Runmin Cong, and Qingming Huang.
\newblock Global context-aware progressive aggregation network for salient
  object detection.
\newblock In {\em AAAI Conference on Artificial Intelligence (AAAI)}, 2020.

\bibitem{RepFinder}
Ming-Ming Cheng, Fang-Lue Zhang, Niloy~J. Mitra, Xiaolei Huang, and Shi-Min Hu.
\newblock Repfinder: Finding approximately repeated scene elements for image
  editing.
\newblock {\em ACM Transactions on Graphics}, 29(4), 2010.

\bibitem{robot_nav}
C. {Craye}, D. {Filliat}, and J. {Goudou}.
\newblock Environment exploration for object-based visual saliency learning.
\newblock In {\em 2016 IEEE International Conference on Robotics and Automation
  (ICRA)}, pages 2303--2309, 2016.

\bibitem{Emeasure}
Deng-Ping Fan, Cheng Gong, Yang Cao, Bo Ren, Ming-Ming Cheng, and Ali Borji.
\newblock Enhanced-alignment measure for binary foreground map evaluation.
\newblock In {\em IJCAI}. AAAI Press, 2018.

\bibitem{AFNet}
Mengyang Feng, Huchuan Lu, and Errui Ding.
\newblock Attentive feedback network for boundary-aware salient object
  detection.
\newblock In {\em Computer Vision and Pattern Recognition (CVPR), 2013 IEEE
  Conference on}, 2019.

\bibitem{obj_retr}
Y. {Gao}, M. {Wang}, D. {Tao}, R. {Ji}, and Q. {Dai}.
\newblock 3-d object retrieval and recognition with hypergraph analysis.
\newblock {\em IEEE Transactions on Image Processing}, 21(9):4290--4303, 2012.

\bibitem{glorot}
Xavier Glorot and Yoshua Bengio.
\newblock Understanding the difficulty of training deep feedforward neural
  networks.
\newblock In {\em Thirteenth International Conference on Artificial
  Intelligence and Statistics}, 2010.

\bibitem{HKU-IS}
Li Guanbin and Yu Yizhou.
\newblock Visual saliency based on multiscale deep features.
\newblock In {\em IEEE/CVF Conference on Computer Vision and Pattern
  Recognition (CVPR)}, 2015.

\bibitem{ResNet}
Kaiming He, Xiangyu Zhang, Shaoqing Ren, and Jian Sun.
\newblock Deep residual learning for image recognition.
\newblock In {\em IEEE Conference on Computer Vision and Pattern Recognition},
  2016.

\bibitem{old_sod1}
L. {Itti}, C. {Koch}, and E. {Niebur}.
\newblock A model of saliency-based visual attention for rapid scene analysis.
\newblock {\em IEEE Transactions on Pattern Analysis and Machine Intelligence},
  20(11):1254--1259, 1998.

\bibitem{EGNet}
Zhao Jia-Xing, Liu Jiang-Jiang, Fan Deng-Ping, Cao Yang, Yang Ju-Feng, and
  Cheng Ming-Ming.
\newblock Egnet: Edge guidance network for salient object detection.
\newblock In {\em International Conference on Computer Vision (ICCV)}, 2019.

\bibitem{PoolNet}
Jiang-Jiang Liu, Qibin Hou, Ming-Ming Cheng, Jiashi Feng, and Jianmin Jiang.
\newblock A simple pooling-based design for real-time salient object detection.
\newblock In {\em IEEE Conference on Computer Vision and Pattern Recognition},
  2019.

\bibitem{old_sod2}
T. {Liu}, J. {Sun}, N. {Zheng}, X. {Tang}, and H. {Shum}.
\newblock Learning to detect a salient object.
\newblock In {\em 2007 IEEE Conference on Computer Vision and Pattern
  Recognition}, pages 1--8, 2007.

\bibitem{sal_track}
V. {Mahadevan} and N. {Vasconcelos}.
\newblock Saliency-based discriminant tracking.
\newblock In {\em 2009 IEEE Conference on Computer Vision and Pattern
  Recognition}, page 1007:1013, 2009.

\bibitem{PiCANet}
Liu Nian, Han Junwei, and Yang Ming-Hsuan.
\newblock Picanet: Learning pixel-wise contextual attention for saliency
  detection.
\newblock In {\em Computer Vision and Pattern Recognition (CVPR)}, 2018.

\bibitem{MINet-CVPR2020}
Youwei Pang, Xiaoqi Zhao, Lihe Zhang, and Huchuan Lu.
\newblock Multi-scale interactive network for salient object detection.
\newblock In {\em CVPR}, June 2020.

\bibitem{old_sod3}
F. {Perazzi}, P. {Krähenbühl}, Y. {Pritch}, and A. {Hornung}.
\newblock Saliency filters: Contrast based filtering for salient region
  detection.
\newblock In {\em 2012 IEEE Conference on Computer Vision and Pattern
  Recognition}, pages 733--740, 2012.

\bibitem{BASNet}
Xuebin Qin, Zichen Zhang, Chenyang Huang, Chao Gao, Masood Dehghan, and Martin
  Jagersand.
\newblock Basnet: Boundary-aware salient object detection.
\newblock In {\em The IEEE Conference on Computer Vision and Pattern
  Recognition (CVPR)}, June 2019.

\bibitem{ECSSD}
Yan Qiong, Xu Li, Shi Jianping, and Jiaya Jia.
\newblock Hierarchical saliency detection.
\newblock In {\em IEEE/CVF Conference on Computer Vision and Pattern
  Recognition (CVPR)}, 2013.

\bibitem{DGRL}
Wang Tiantian, Zhang Lihe, Wang Shuo, Lu Huchuan, Yang Gang, Ruan Xiang, and
  Borji Ali.
\newblock Detect globally, refine locally: A novel approach to saliency
  detection.
\newblock In {\em Computer Vision and Pattern Recognition (CVPR)}, 2018.

\bibitem{DUTs}
Lijun Wang, Huchuan Lu, Yifan Wang, Mengyang Feng, Dong Wang, Baocai Yin, and
  Xiang Ruan.
\newblock Learning to detect salient objects with image-level supervision.
\newblock In {\em Computer Vision and Pattern Recognition (CVPR), 2013 IEEE
  Conference on}, 2017.

\bibitem{F3Net}
Jun Wei, Shuhui Wang, and Qingming Huang.
\newblock F3net: Fusion, feedback and focus for salient object detection.
\newblock In {\em AAAI Conference on Artificial Intelligence (AAAI)}, 2020.

\bibitem{LDF}
Jun Wei, Shuhui Wang, Zhe Wu, Chi Su, Qingming Huang, and Qi Tian.
\newblock Label decoupling framework for salient object detection.
\newblock In {\em IEEE/CVF Conference on Computer Vision and Pattern
  Recognition (CVPR)}, June 2020.

\bibitem{CPD}
Zhe Wu, Li Su, and Qingming Huang.
\newblock Cascaded partial decoder for fast and accurate salient object
  detection.
\newblock In {\em The IEEE Conference on Computer Vision and Pattern
  Recognition (CVPR)}, June 2019.

\bibitem{SCRN}
Zhe Wu, Li Su, and Qingming Huang.
\newblock Stacked cross refinement network for edge-aware salient object
  detection.
\newblock In {\em The IEEE International Conference on Computer Vision (ICCV)},
  October 2019.

\bibitem{old_sod4}
Q. {Yan}, L. {Xu}, J. {Shi}, and J. {Jia}.
\newblock Hierarchical saliency detection.
\newblock In {\em 2013 IEEE Conference on Computer Vision and Pattern
  Recognition}, pages 1155--1162, 2013.

\bibitem{DUT-OMRON}
Chuan Yang, Lihe Zhang, Huchuan Ruan~Xiang Lu, and Ming-Hsuan Yang.
\newblock Saliency detection via graph-based manifold ranking.
\newblock In {\em Computer Vision and Pattern Recognition (CVPR), 2013 IEEE
  Conference on}, 2013.

\bibitem{PASCAL-S}
Li Yin, Hou Xiaodi, Koch Christof, M.~Rehg James, and L.~Yuille Alan.
\newblock The secrets of salient object segmentation.
\newblock In {\em IEEE/CVF Conference on Computer Vision and Pattern
  Recognition (CVPR)}, 2014.

\end{thebibliography}
}
\end{document}